**Modified-Improved Fitness Dependent Optimizer for Complex and Engineering Problems**

**Hozan K. Hamarashid[1], Bryar A. Hassan[2,3], Tarik A. Rashid[2]**

[1]Information Technology Department, Computer Science Institute, Sulaimani Polytechnic University, Sulaimani, Iraq

[2]Computer Science and Engineering Department, University of Kurdistan Hewler, Erbil, Iraq

[3]Department of Computer Science, College of Science, Charmo University, 46023 Chamchamal/Sulaimani, KRI, Iraq

Email (Corresponding): bryar.ahmad@ukh.edu.krd

**Abstract**

Fitness dependent optimizer (FDO) is considered one of the novel swarm intelligent algorithms. Recently, FDO has been enhanced several times to improve its capability. One of the improvements is called improved FDO (IFDO). However, according to the research findings, the variants of FDO are constrained by two primary limitations that have been identified. Firstly, if the number of agents employed falls below five, it significantly diminishes the algorithm's precision. Secondly, the efficacy of FDO is intricately tied to the quantity of search agents utilized. To overcome these limitations, this study proposes a modified version of IFDO, called M-IFDO. The enhancement is conducted by updating the location of the scout bee to the IFDO to move the scout bees to achieve better performance and optimal solutions. More specifically, two parameters in IFDO, which are alignment and cohesion, are removed. Instead, the Lambda parameter is replaced in the place of alignment and cohesion. To verify the performance of the newly introduced algorithm, M-IFDO is tested on 19 basic benchmark functions, 10 IEEE Congress of Evolutionary Computation (CEC-C06 2019), and five real-world problems. M-IFDO is compared against five state-of-the-art algorithms: Improved Fitness Dependent Optimizer (IFDO), Improving Multi-Objective Differential Evolution algorithm (IMODE), Hybrid Sampling Evolution Strategy (HSES), Linear Success-History based Parameter Adaptation for Differential Evolution (LSHADE) and CMA-ES Integrated with an Occasional Restart Strategy and Increasing Population Size and An Iterative Local Search (NBIPOP-aCMAES). The verification criteria are based on how well the algorithm reaches convergence, memory usage, and statistical results. The results show that M-IFDO surpasses its competitors in several cases on the benchmark functions and five real-world problems.

## 1. Introduction

Searching to discover the most effective solution was interesting from the development of the computer due to it is important to reach the target with the best outcome. However, there were many various ways to discover optimal solutions. For this purpose, in 1945, Alan Turning used a search optimization algorithm for the first time during the Second World War [1]. Thus, solving complicated problems through traditional techniques relied on logic or mathematical programming developed [2]. Optimization is an aspect or method in which many algorithms were made and developed to control its limitations and difficulties. In optimization, a method's best value or solution will be achieved by searching for a parameter. The list of available values or solutions is considered a list of possible solutions, but one is the best value or solution [3]. Optimization is a process, an act, or a method of building things, for example, a system or a design, or even making a perfect decision. It could be effective or functional, as much as possible, particularly in the procedures of mathematical equations, for instance, discovering the maximum value of the included functions [4]. There are optimization difficulties, minimizing, and maximizing of several functions that are related to several sets, mostly, it is demonstrating a range of options that are available and provided in a confident position. These functions acknowledge comparing the various options or choices to indicate which option or choice would be the best. Several examples of optimization applications are maximum profit, minimum cost, and error. So, dealing with and solving optimization difficulties is the main aim of conceiving optimization algorithms. Optimization algorithms are divided into two categories: deterministic and stochastic algorithms. The deterministic type constructs a similar set of solutions in case a similar starting point is utilized to start the iteration. It is because of utilizing tendency. In contrast, stochastic-type algorithms commonly construct various solutions with similar values without using tendency. Consequently, there are very small differences in the ending consequences. Thus, a very similar best value matched the determined accuracy. The stochastic type is also divided into two groups which are heuristic and metaheuristic [4]. Heuristic techniques use the error and preliminary to search for a solution. Obtaining a solution is anticipated to take a reasonable amount of time. In addition, it uses various techniques in randomization methods and local investigations or explorations [5]. Therefore, some researchers have developed heuristic techniques and reconstructed them into metaheuristic methods, so they perform much better than heuristic methods. Thus, the meta prefix that means higher is combined with heuristic algorithms [6]. Consequently, the entire meta-heuristic techniques try to stabilize randomization and local exploration [7]. Today, the problems of the real world have become more complex, and it is difficult to find convenient solutions if we consider space, time, and cost. To solve these difficulties, feasible low-cost, and fast methods are crucial. Therefore, researchers discovered animal behaviors and natural circumstances to find and tackle these difficulties. For example, how ants discover or select their paths, and how gravity works. Thus, the name nature was determined for algorithms that were inspired by nature and named nature-inspired algorithms. For the first time, Michigan University started to invent these algorithms in 1960, and the GA was published by Holland [8].

Originally, Jaza and Tarik introduced the FDO [9]. To examine the reproductive strategy and mass decision-making behaviors of a bee swarm, which imitates the collective behavior of a bee swarm in locating new colonies. The algorithm proposed is partially derived from the Particle Swarm Optimization (PSO) technique. Therefore, the agent position update method imitates the PSO algorithm, but with certain differences. The advantages of using the fitness weight (fw) technique in FDO include its ability to explore and exploit effectively, its tendency to be limited to local optima, its faster convergence time, and its superior performance in solving certain real-world issues compared to other metaheuristic algorithms. The FDO method demonstrated very competitive performance when compared to various metaheuristic algorithms. Furthermore, it has been successfully applied to numerous real-world applications, resulting in significant improvements.

The FDO algorithm demonstrates superior competitiveness when compared to other metaheuristic algorithms. It has been successfully utilized to address various real-world problems, including aperiodic antenna array designs (AAAD), frequency-modulated sound waves (FM) [9], rainfall data [10], one-dimensional bin packing problem [11], pressure vessel design (PVD), and task assignment problem [12].

Metaheuristic algorithms encounter challenges when it comes to locating the optimal answer. Metaheuristic algorithms possess the capacity to tackle complex optimization issues [13]. However, metaheuristic algorithms that rely on stochastic mechanisms possess the capability to achieve global optima and surpass local optima [14]. Metaheuristic algorithms rely on the incorporation of swarm interaction to provide efficient exploration [15] and a balance between exploration and exploitation [16].

FDO, while a promising optimization technique, faces several challenges in its application that urge scholars to modify it. Premature convergence, as discussed by [17], remains a significant issue, restricting FDO from thoroughly exploring the solution space and potentially settling for suboptimal solutions. Moreover, sensitivity to parameter settings can significantly impact FDO's performance, leading to inefficient convergence or trapping in local optima. Achieving an optimal balance between exploration and exploitation, a critical aspect of optimization, is often challenging for FDO, potentially limiting its efficiency in navigating complex landscapes. Scalability issues, especially in handling high-dimensional problems, pose concerns regarding FDO's applicability to larger-scale optimization tasks. Additionally, the algorithm's robustness in noisy environments and the need for comprehensive benchmarking and validation further contribute to the challenges faced by FDO. These limitations necessitate continued research efforts to enhance FDO's performance, robustness, and versatility across diverse optimization scenarios. On that basis, the primary objective of this study is to modify IFDO to address the problems of FDO's variants, including its limited capacity to be effectively utilized and its slow convergence speed. Ultimately, FDO is employed to address practical and tangible real-world scenarios. Also, the contribution of this algorithm is to enhance the capability of the IFDO. To develop the IFDO, the M-IFDO computes the Lambda instead of alignment and cohesion and then utilizes it with pace as the IFDO renews its position. In addition, weights are determined in IFDO as a weight factor ($wf$). IFDO utilizes and performs $wf$ randomly between (0, 1), and then when a better fitness weight value is obtained, the range will be minimized. The M-IFDO and its Convergence method for discovering the optimal value which is represented in this paper.

The remainder of this paper is organized as follows. The basics of FDO is introduced in Section 2. Section 3 discusses the details of related work. Section 4 describes the proposed algorithm, while Section 5 explains the experimental methodology of the study, including testing functions, parameter tuning, and five real-world applications of FDO. The results of the study are discussed in Section 6. The conclusion of the research is drawn in Section 7.

## 2. The Fitness Dependent Optimizer

A recently developed swarm intelligence algorithm called FDO was put forth by [9]. It is based on the characteristics of bees swarming during the process of procreation to locate better hives. The position update mechanism of the FDO slightly imitates the PSO algorithm. Finding the best hives is the major goal of the search agents, which are scout bees, after randomly initializing their population in the search space. If the previous solution is not superior to the present one, the scout bees disregard it [11]. Scout bees alter their position depending on the prior best position while ignoring the low solution if it does not become better. The population of scout bees is denoted by the notation *Xi* (i= 1, 2, . . ., n) [10][16]. Scout bees employ two techniques to look for objects inside the search area: *fw* and scout bee movement procedure. To get a better result, the scout bee in this algorithm adjusts its present location based on speed. Equation (1) is used to compute the scout bee movement.

$$X_{i+1} = X_{i,t} + pace \tag{1}$$

Where:

X: artificial search agent (scout bee),

I: current search agent,

t: the current iteration.

The pace can be used to determine the artificial search agent's movement rate and direction. The value of the *fw* and the randomization techniques also affect pace. The computation of *fw* based on best fitness and current fitness is shown in Equation (2).

$$fw = \left[\frac{X^*_{i,t\,fitness}}{X_{i,t\,fitness}}\right] - wf \tag{2}$$

Where:

$X^*_{i,t\,fitness}$ is the fitness function of the global best solution,

$X_{i,t\,fitness}$ is the current best solutions of the scout bee,

*wf* is the weight factor, and it has a range of [0,1].

Furthermore, the Levy flight method is the basis for the random number generator r used by FDO. This random number ranges from -1 to 1. Also, dependent on the *fw* and *r*, FDO offers three alternative conditions for computing pace value. The pace value conditions are shown in equation (3). However, when wf is equal to zero, *wf* has no impact on Equation (2). Calculating the pace value involves multiplying the current position by *r*. As can be presented in Equation (3), if *wf* = 1. Then, if *wf* is equal to zero, the pace can be calculated by multiplying $distance_{bestbee}$ with r. Thereafter, if (1 > *fw* > 0) and *r* < 0, pace equals pace* - 1. Meanwhile, in otherwise conditions Equation (3) shows how to calculate the pace.

$$pace = \begin{cases} X^*_{i,t}r & if\ fw = 1 \\ distance^*_{best\ bee}r & if\ fw = 0 \\ pace^* - 1 & if\ fw > 0\ and\ fw < 1\ and\ r < 0 \\ distance^*_{best\ bee}fw & if\ fw > 0\ and\ fw < 1\ and\ r \geq 0 \end{cases} \quad (3)$$

where $distance_{best}$ bee denotes the variation in the current agent from the best agent. Hence, it can be calculated by Equation (4).

$$distance_{best\ bee} = X^* - X_{i,t} \quad (4)$$

Figure 1 presents the details of FDO algorithm.

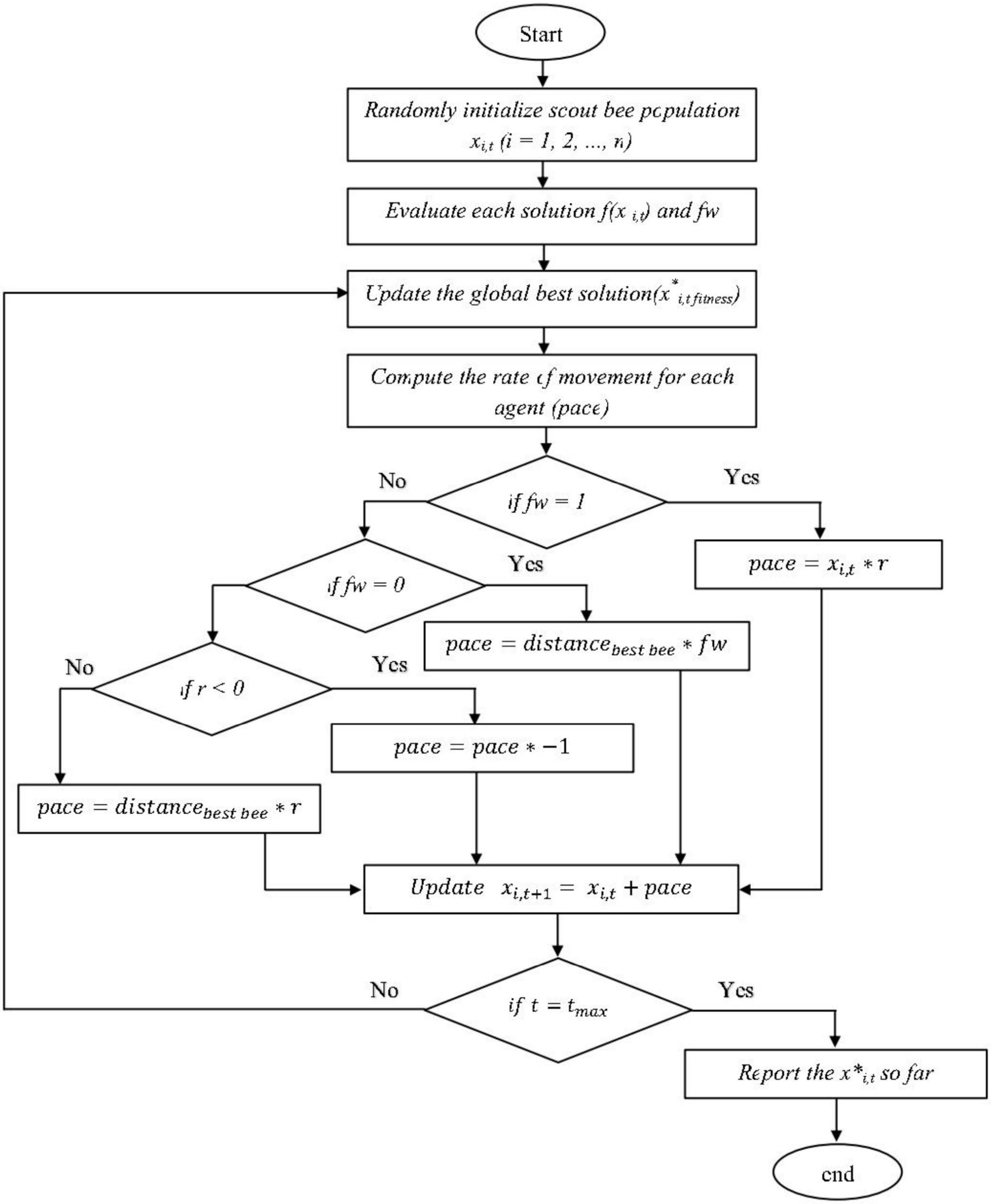

**Figure 1: Flowchart of the FDO (extracted from** [11]**)**

When it comes to finding the best solution and accelerating convergence, metaheuristic algorithms are both effective and constrained. Due to the presence of numerous randomized parameters, including *fw*, *wf*, pace, update Equation, optimum updating solution, and Levy flight, FDO has restrictions. Subsequently, Due to a convergence problem with the FDO, setting *wf* to zero results in low convergence. Another issue with FDO is that it employs the fitness of the best agent, current fitness, and *wf*, which results in an incorrect balance between exploration and exploitation [8]. The disparity between the two stages is also caused by tempo based on randomness. FDO's performance is bad because of solution refinement. This issue resulted from using a set value for *wf* and employing pacing equations to achieve poor outcomes [9]. According to literature-based research, FDO performs poorly compared to other competitive algorithms.

## 3. Related Work

PSO and ant colony optimization ACO are the most common algorithms that utilize swarm intelligence algorithms which are projected by [18][19]. A group of birds' behaviors activates PSO in looking for food. Besides, ACO is activated by the ant's nature, which can influence earlier paths in its mind [20][21]. Because of these behaviors, PSO authors thought that optimization difficulties would be limited. In addition, PSO is significant for several other algorithms. Different significant intelligence swarms have been proposed in the past 20 years. For instance, Differential Evolution DE was developed in 1997 [22]. It was better than GA in many aspects and depended on vector based. Later, Artificial Bee Colony was discovered in 2005 [23]. The Firefly Algorithm (FA) was suggested in 2009 as a swarm intelligence-based approach and has been demonstrated to be successful in resolving nonlinear optimization issues, particularly multimodal issues where the goal landscape may contain several maxima or minima [24]. After that, in 2009, the author of FA proposed the CS algorithm [25]. In 2010, the bat-inspired algorithm was proposed [26]. Another algorithm was developed based on the plant growth process called the Artificial Plant Optimization Algorithm (APOA) in 2013 [27]. In addition, in 2014, [28] proposed an animal migration optimization algorithm (AMO). It depended on swarm movement behavior. Continuously, in 2015, 2016, and 2017, based on attracting food and enemy evasion, DA, WOA, and SSA algorithms were proposed, respectively [29]. So, several algorithms were developed, and lastly, in 2019 FDO algorithm was invented [9]. In the FDO algorithm, the behavior of bee swarms is looked at during reproduction. Besides, it imitates swarm activities. A partial part of this algorithm is discovering applicable solutions to various available solutions. Consequently, no method can address all the optimization problems. In contrast, several algorithms were utilized to solve specific optimization problems. Thus, discovering better developed or modified optimization algorithms still opens difficult cases [30].

### 3.1. Past Works on FDOs

Table (1) summarizes the modifications and hybridization of FDO.

**Table (1): The modifications and hybridization of FDO**

| **Author name, Year** | **Refe renc es** | **Brief of existing algorithm** | **Benchmarks used** | **Results** | **Applications** | **Limitation** |
|---|---|---|---|---|---|---|
| J.M. Abdullah, T. Ahmed, 2019 | [9] | FDO is considered a particle swarm optimization (PSO)--based algorithm that updates the search agent position by adding velocity (pace). | 19 classical benchmark functions and CEC-C06 2019 Competition | Outperforms PSO, GA, dragonfly algorithm (DA), WOA, SSA | Aperiodic antenna array designs and frequency-modulated sound waves | Using a small number of agents (below five) would notably decrease the accuracy of the algorithm. |
| D.A. Muhammed et al., 2020 | [8] | IFDO calculates the alignment and cohesion and then uses these behaviors with the pace at which the FDO updates its position. | 19 classical benchmark functions and CEC-C06 2019 Competition | Outperforms FDO | Aperiodic antenna array designs and pedestrian evacuation model | The limitation of the FDO algorithm is that its performance depends on the number of search agents used. The accuracy of the algorithm suffers noticeably when it uses fewer than five search agents. Conversely, using a large number of search agents enhances the accuracy and rate with more space and time. |
| J.M. Abdullah et al., 2023 | [31] | Multi-objective FDO (MOFDO) is a multi-objective variant of the recently introduced fitness fitness-dependent optimizer (FDO). | classical ZDT functions and CEC-C06 2019 Competition | Outperforms multi-objective PSO, non-dominated sorting genetic algorithm third improvement (NSGA-III), and multi-objective DA | Welded beam design problems | Not addressed. |
| D.S. Abdul-Minaam et. al., 2020 | [32] | Adaptive FDO is an adaptive procedure using FDO to solve the bin packing problem (BPP). | 30 instances obtained from benchmark datasets | outperforms the PSO, crow search algorithm (CSA) and Jaya algorithm by 16%, 17%, and 11%, respectively | One-dimensional bin packing problem | Not addressed. |
| D.Kh. Abbas et al., 2022 | [33] | FDO and multilayer perceptron neural networks (FDO-MLP) combine FDO with MLP (codename FDO-MLP) for optimizing weights and biases to predict outcomes of students. | Student dataset | Superior to back-propagation algorithm (BP), Grey Wolf Optimizer (GWO) combined with MLP (GWO-MLP), modified GWO combined with MLP (MGWO-MLP), GWO with cascade MLP (GWO-CMLP), and modified GWO with cascade MLP (MGWO-CMLP). | Classification of student outcomes | Not addressed. |
| P.C. Chiu et al., 2021 | [34] | The sine cosine algorithm and FDO (SC-FDO) propose a novel hybrid of the sine cosine algorithm and fitness-dependent optimizer (SC-FDO) for updating the velocity (pace) using the sine cosine scheme. | 19 classical benchmark functions and CEC-C06 2019 Competition | Superior to six existing nature-inspired algorithms, such as FDO, IFDO, SCA, WOA, PSO, and BOA. | Missing weather data imputation | Not addressed. |
| M.T. Abdulkhaleq et al., 2023 | [35] | FDO_MLP and FDO_CMLP are proposed to diagnose to classify Covid-19 patients. | Textual clinical dataset | FDO_MLP and FDO_CMLP outperform the other models | COVID-19 | Not addressed. |
| H.M. Mohammed and T.A. Rashid, 2021 | [36] | This algorithm aims at improving the performance of FDO; thus, the chaotic theory is used inside FDO to propose chaotic FDO (CFDO). | 10 benchmark functions of CEC2019, CEC2013 and CEC2005 | CFDO is superior to GA, FDO, and CSO | Pressure vessel design | Not addressed. |
| J.F. Salih et al., 2022 | [37] | To fix FDO, MFDO is proposed. To improve FDO performance, MFDO optimised the weight factor range between 0 and 0.2 for fitness weight. Second, updating fitness weight and pace (bee speed) with sine cardinal mathematical function. | 19 classical benchmark functions and CEC-C06 2019 Competition | MFDO achieves significant performance compared to Grey Wolf Optimization (GWO), Chimp Optimization Algorithm (ChOA), Genetic Algorithm (GA), and Butterfly Optimization Algorithm (BOA). | Welded beam design, pressure vessel design, and spring design problem | Not addressed. |
| M. Zivkovic et al., 2022 | [38] | FDO-XGBoost is proposed for tackling intrusion detection systems face. | NSL-KDD benchmark dataset | FDO-XGBoost significantly outperforms other approaches in terms of accuracy, having precision and recall with average values of 0.82 and 0.77, respectively | Network intrusion detection | Not addressed. |
| B.H. Tahir et al., 2022 | [39] | FDO improved in a variant of the FDO motivated by scout bees in the hive exploring the process of seeking food | standard 24-unit system | IFDO obtains 7.94E−12, the lowest transmission. | Economic Load Dispatch Problem | Fine-tuning of FDO parameters in combination with ELD constraint and their hyperparameter tuning is suggested. |
| | | from a pool of suitable options. | | | | |
| A. Daraz et al., 2020 | [40] | I-FDO is proposed to optimize the parameters of the controller. | FOI-PD controller | I-FDO performs superior in terms of less Overshoot (Os), Settling time (Ts), and Undershoot (Us). | Automatic generation control of multi-source interconnected power system | Not addressed. |
| G.F. Laghari et al., 2023 | [41] | GA and FDO are hybridized to solve dynamical system Nonlinear Optimal Control Problems (NOCPs) with the best numeric answer. | Three real-world NOCPs, including Van der Pol (VDP) oscillator problem, Chemical Reactor Problem (CRP), and Continuous Stirred-Tank Chemical Reactor Problem (CSTCRP). | The proposed technique generates a better solution and surpasses the recently represented methods in the literature. | Optimal control problems of nonlinear dynamical systems | Not addressed. |
| K.G. Dhal et al., 2023 | [42] | A chaotic Fitness-Dependent Quasi-Reflected Aquila Optimizer (CFDQRAO) is | images of blood pathology | CFDQRAO proves itself better than other strategies in terms of visual analysis and the | segmenting white blood cells | Not addressed. |
| | | proposed to prevail over the local trapping problem. | | values of segmentation quality parameters. | | |

### 3.2. The Need for Modifying FDO Variants

In 2020 FDO was improved by [8] and named IFDO. The IFDO algorithm added alignment and cohesion of scout bee's behavior to the pace. In this paper, however, most of the IFDO characteristics remain the same. Still, M-IFDO is proposed by removing alignment and cohesion parameters which need more time to compute and replacing them with Lambda parameters. In addition, in most cases, M-IFDO is better than IFDO. In the proposed algorithm M-IFDO weight factor is utilized when improved fitness weight is achieved. According to the research findings, the research gap of FDO is constrained by two primary limitations that have been identified.

- Firstly, if the number of agents employed falls below five, it significantly diminishes the algorithm's precision.
- Secondly, the efficacy of FDO is intricately tied to the quantity of search agents utilized. When the number of search agents drops below five, the algorithm's accuracy experiences a noticeable decline. Conversely, employing a larger cohort of search agents amplifies both accuracy and efficiency, albeit at the cost of increased computational resources and time. Consequently, in response to these identified drawbacks of FDO, modifications have been made to IFDO aimed at addressing these specific limitations.

It needs to mention that FDO has not hybridized with the other competitive algorithms such as particle swarm optimization (PSO) [43], moth flame optimization algorithm (MFO) [44], slime mold algorithm (SMA) [45] [46], backtracking search optimization algorithm (BSA) [47][48], symbiotic organisms search algorithm (SOS) [49][50] and differential evolution [51].

## 4. The Proposed Algorithm

The M-IFDO algorithm is a modification of the IFDO algorithm. This is an evolutionary algorithm that was developed by [8]. The main concept of the proposed algorithm relies on discovering or looking for several or many conceivable hives to achieve a new appropriate hive. There are two aspects to renewing the scout bee position. This is modified by omitting and replacing parameters. In addition, the randomization of the weight factor between (0,1) remains the same. In the IFDO algorithm, movement alignment and cohesion were added to compute the divergence. This is modified and improved by omitting alignment and cohesion parameters and replacing them with Lambda. Besides, Lambda added to the pace. Pace is utilized to achieve an appropriate solution to the scout bee for the current position in discovering or looking for a new position as shown in Equation (1).

So, in the M-IFDO, the alignment and cohesion are removed, and Lambda is added to the pace in the equation as represented in Equation (5).

$$Xi,t+1 = Xi,t + pace + Lambda \quad (5)$$

To illustrate the equation, *i* represents the current search agent or artificial scout bee. *t* represents the current iteration. Where the pace represents the ratio of artificial bee orientation and movement. Lastly, Lambda is a parameter added to the pace, which has a value of 0.1; by this, a better solution and results are obtained compared to several well-known algorithms, especially FDO and IFDO algorithms. In addition, it improves performance because it needs less computation time, such as computing alignment and cohesion, which needs more time. Moreover, the computation of space is omitted. The pace is computed as shown in Equations (6), (7), and (8).

$$fw \ = \ 1 \ or \ fw = 0 \ or \ xi,t \ fitness = 0, pace = \ xi,t * r \tag{6}$$

$$fw > 0 \ and \ fw < 1 \{ r < 0, pace = (xi,t - \ xi,t*) * fw * \ -1 \tag{7}$$

$$r \geq 0, pace = (xi,t - \ xi,t*) * fw \tag{8}$$

M-IFDO also has time complexity (d*p + COF*p), in each iteration. *d* represents the dimension of the problem, where *p* is the population size, and the last COF is the objective function cost. The weight factor M-IFDO is still the same as IFDO and randomized between the range of (0,1) as shown in Equation (9).

$$fw = |xi,t \ fitness * \ xi,t \ fitnees \ | \tag{9}$$

In equation (10), we can discover the fitness weight, after that, it can be checked if its value is less than or equal to the produced weight factor. If it is less than the value will be ignored in handling the fitness weight. Alternatively, it takes part in the activity of handling the fitness weight as shown in Equation (10).

$$fw = fw - wf \tag{10}$$

To find fitness weight that is ignored by avoiding $wf$ in most of the cases. However, $wf$ sensibly takes part in activity in most cases. So, in the entire iteration for each scout, the weight factor is randomly set. Besides, when a better and more recent solution is approved this means that in the range of (0, $wf$), the new $wf$ is produced. The pseudo-code of M-IFDO is presented in Algorithm (1).

**Algorithm (1): The pseudo-code of M-IFDO**

```
Initialize scout bee population Xt, (i = 1, 2, ..., n)
Generate random weight factor (wf) in [0, 1] range
while iteration (t) limit is not reached
 for each artificial scout bee Xt,i
  find best artificial scout bee xt,i*
  generate random-walk r in [-1, 1] range
  if( Xt,i fitness == 0) (avoid dividing by zero)
   fitness weight = 0
  else
   calculate fitness weight, Equation (10)
     if(fitness weight > wf)
  calculate fitness weight, Equation (11)
  end if
   end if
   if (fitness weight = 1 or fitness weight = 0)
    calculate pace using Equation (7)
   else
     if (random number >= 0)
       calculate pace using Equation (9)
    else
       calculate pace using Equation (8)
     end if
   end if
  Lambda = 0.1
  calculate Xt+1,i, Equation (6)
  if( Xt+1,i fitness<Xt,i fitness)
     move accepted and pace saved
 generate new wf in [0, wf]
  else
   calculate Xt+1,i, Equation (6)
        ... with previous pace
  if (Xt+1,i fitness<Xt,i fitness)
   move accepted and save pace
        generate new wf in [0, wf]
  else
     maintain current position (don't move)
  end if
    end if
  end for
 end while
```

Furthermore, M-IFDO's mathematical convergence analysis typically involves assessing several key aspects related to its convergence behavior. While an exact mathematical convergence proof for M-IFDO might vary depending on the specific implementation or variant of the algorithm, convergence analysis generally revolves around these fundamental concepts:

**1. Convergence Criterion:** Defining a convergence criterion or condition that determines when the algorithm has sufficiently converged. This could involve assessing the difference between successive objective function values or the distance between the current solution and the known optimal solution as presented in Equation (11)

$$\left|f(x_t) - f_{optimal}\right| < \epsilon \qquad (11)$$

Here, $f(x_t)$ represents the objective function value at iteration $t, f_{optimal}$ is the known optimal or target objective function value, and $\epsilon$ is a predefined tolerance level.

**2. Rate of convergence:** Assessing the rate at which the algorithm approaches the optimal solution. This might involve analyzing the change in objective function values across iterations to understand how quickly the algorithm converges as shown in Equation (12).

$$|f(x_t) - f(x_{t-1})| \quad (12)$$

This equation measures the difference between the objective function values at consecutive iterations *t* and *t-1*.

**3. Convergence behavior:** Observing the behavior of the algorithm's solutions across iterations. This includes plotting fitness values against iterations to identify convergence trends, oscillations, or stagnation in the search process.

**4. Convergence analysis techniques:** Employing mathematical tools and techniques, such as convergence proofs or analyses under certain assumptions or conditions, to understand M-IFDO's theoretical convergence properties. These techniques might involve proving convergence to a global optimum or characterizing convergence behavior in specific problem scenarios.

The specific mathematical convergence analysis of M-IFDO would depend on its formulation, parameters, and the problem being solved. Conducting a detailed mathematical analysis often involves rigorous proofs, empirical studies, and experimentation across various problem instances to understand the algorithm's convergence behavior and properties. The specific mathematical details for the convergence analysis of M-IFDO might be found in research papers or studies dedicated to analyzing its convergence properties.

Additionally, In the M-FDO, constraints are managed through a combination of penalty methods and repair algorithms.

- Firstly, penalty methods are integrated into the fitness calculation to penalize solutions that violate constraints. This penalty incentivizes the algorithm to prioritize feasible solutions during selection and optimization.
- Secondly, repair algorithms are employed to adjust infeasible solutions, transforming them into feasible ones while preserving their essential characteristics.
- Finally, by combining these approaches, M-FDO effectively ensures that constraints are adhered to throughout the optimization process, maintaining the integrity of the solutions while exploring the solution space.

## 5. Experimental Methodology

In this section, we present the proposed experimental methodology of the research, including testing functions, parameter tuning, and a case study to apply M-IFDO to a real-world problem. Figure 2 depicts the overall steps of our experimental method.

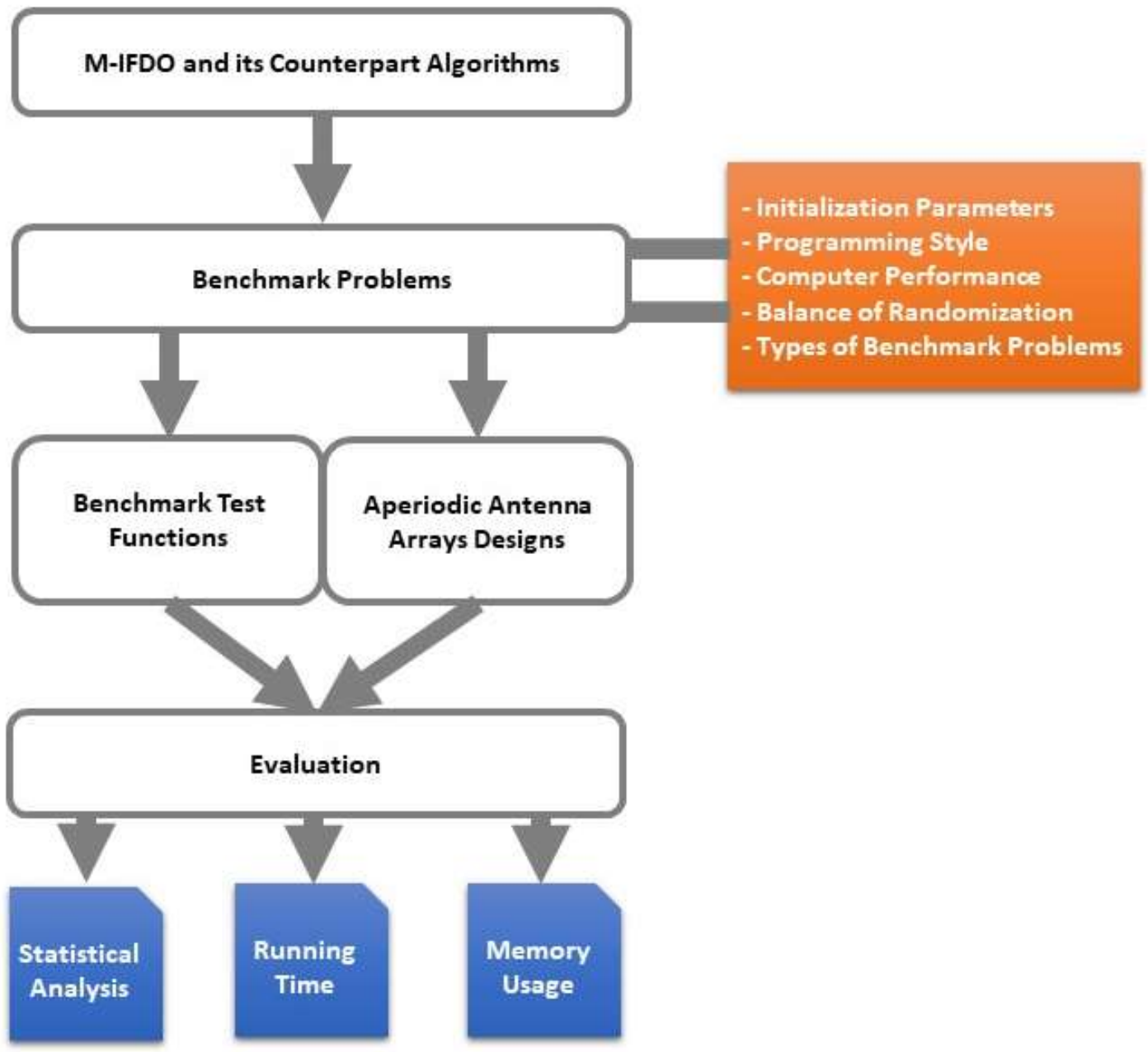

**Figure 2: The experimental methodology of the study**

### 5.1. Testing Functions

The standard and CEC-C06 2019 benchmark testing functions are employed to assess the performance of M-IFDO.

Referencing the standard benchmark functions, three categories are utilized, which are unimodal, multimodal, and composite [52][53]. The unimodal function is used to validate exploration and convergence levels to conclude a single optimum. In contrast, the multimodal test function contains many optimal solutions. That means it validates the evasion of local optima and the level of exploration. According to the existence of many solutions, most of them are local, and one of them is global. Lastly, different search areas can contain various forms and many local optima in composite test functions. In general, composite test functions are transported, changed, and biased adoption of other test functions. So, real-world search field problems could be determined by this kind of test function.

A further assessment was conducted on the M-IFDO by testing it on the CEC standard functions containing ten groups of functions. CEC normal functions for single objective problem optimization were developed by [54]. These test functions are considered for a yearly optimization competition. They analyze large-scale optimization algorithms. CEC01–CEC03 have different dimensions. However, CEC04 through CEC10 functions are shifted and rotated to solve 10-dimensional minimization problems in the range [-100, 100]. All scalable CEC functions have a global optimum at point 1.

### 5.2. Parameter Tuning

Because the solution of the optimization algorithm can differ between runs, the entire algorithm is examined 30 times with 10 dimensions, 30 scout bees, and 500 iterations. [53] illustrates parameters for PSO, DA, and GA. Wf in M-IFDO is randomly set in between (0, 1). In contrast, the value is altered when a better value or solution is found. Table (2) represents the setting parameters of M-IFDO.

**Table (2): Parameter tuning of M-IFDO**

| Parameter | Parameter value |
|---|---|
| Number of runs | 30 |
| Number of iterations | 500 |
| Dimensions | 10 |
| Scout bees | 30 |
| Wf | [0, 1] |

Furthermore, to test M-IFDO, MATLAB programming language on an MSI prestige personal computer is used with hardware specification of 32 RAM and Core i7 Processor and Windows 11 as a software operator. Meanwhile, the M-IFDO algorithm performance is assessed and compared with several algorithms, IFDO, FDO, DA, GA, PSO, SSA, and WOA. The consequences of the standard test functions and CEC-C06 tests of the various algorithms are taken from IFDO research work [8].

### 5.3. Real-world Problems

In this section, we have briefly presented five real-world problems as case studies applied to M-IFDO and its counterpart algorithms.

### A. Aperiodic Antenna Arrays Designs

In the 1960s, significant improvements in radar and radio astronomy methods derived concentration to Aperiodic Antenna Arrays Designs (AAAD). To present a position in non-uniform arrays, real-number vectors are essential. This is to optimize the feature position to obtain the highest sidelobe level SLL. In addition, the feature boundary position satisfaction is necessary to prevent discordant lobes. More information can be obtained in [55]. Equation (13) illustrates the constraints of this four-dimensional optimization problem.

$$\text{x_i} \in |\text{x_i} - \text{x_j}\,|(0{,}2.25) > \min\{\text{xi}\}\ 0.25\lambda_0 > 0.125\lambda_0.\, \text{i} = 1{,}2{,}3{,}4.\, \text{i} \neq \text{j}. \quad (13)$$

Nevertheless, no feature is larger than 2.0λ0 or smaller than 0.125λ0 because of the restraints and every feature has a boundary between 0 and 2.25 due to the feature 2.25λ0 being stable, and the neighbor features cannot be close to 0.25λ0. The difficulty of the fitness function is shown in Equation (14).

$$f = \{20\ log\ |(\theta)|\} \quad (14)$$

Where:

AF(θ)= Σcos[(cosθ−cosθs)2πxi]4i=1 +cos[(cosθ−cosθs)2.25×2π]

**B. Travelling Salesman Problem**

The Traveling Salesman Problem (TSP) is one of the most well-known combinatorial optimization problems in the field of operations research and computer science [56]. It involves finding the shortest possible route that visits a set of cities exactly once and returns to the starting city. The objective is to minimize the total distance or travel time required to complete the tour. The problem is characterized by its exponential complexity, meaning that the number of possible solutions grows rapidly with the number of cities, making it computationally challenging to solve large problem instances. The TSP has numerous applications in logistics, transportation, and routing optimization, where finding an optimal or near-optimal solution can lead to significant cost savings and efficiency improvements. Various algorithms are employed to tackle the TSP, including exact methods like dynamic programming, and branch-and-bound, as well as heuristic and metaheuristic approaches such as genetic algorithms, simulated annealing, and ant colony optimization. Despite its computational complexity, the TSP remains a fundamental problem in combinatorial optimization, with ongoing research focused on developing efficient solution techniques and addressing real-world variants and applications.

**C. Combinatorial Optimization Problem**

Combinatorial optimization problems (COP) involve finding the best solution from a finite set of possibilities, characterized by discrete decision variables and an objective function subject to constraints [57]. These problems, prevalent in fields like computer science and operations research, pose challenges due to their vast solution spaces, rendering exhaustive searches impractical. Instead, techniques such as heuristic algorithms, integer programming, and metaheuristics are employed to efficiently navigate the solution space and find optimal or near-optimal solutions. Examples include the traveling salesman problem and the knapsack problem, with applications spanning logistics, scheduling, and bioinformatics, where effective solutions can lead to significant cost savings and improved resource allocation.

**D. Job-shop Scheduling**

The Job-shop scheduling problem (JSP) is a classic combinatorial optimization problem in operations research and production management. It involves scheduling a set of jobs on a set of machines, where each job consists of a sequence of operations that must be processed on specific machines in a predefined order [58]. The objective is to determine the optimal schedule that minimizes a certain criterion, such as makespan (the total time required to complete all jobs) or total completion time. The constraints typically include precedence constraints (certain operations must be completed before others can start) and resource constraints (each machine can process only one operation at a time). The Job-shop scheduling problem is NP-hard, meaning that finding an optimal solution becomes increasingly difficult as the problem size grows. Various algorithms and heuristics, such as genetic algorithms, simulated annealing, and branch-and-bound techniques, are employed to find near-optimal solutions efficiently. The problem finds applications in manufacturing, production planning, and scheduling across a range of industries, where efficient job sequencing can lead to improved resource utilization, reduced production time, and increased productivity.

**E. Pressure Vessel Design Problem**

The pressure vessel design problem is a critical engineering challenge in various industries, particularly in the oil and gas, chemical, and aerospace sectors [59]. It involves the optimization of vessel geometry, material selection, and operating conditions to ensure structural integrity and safety while minimizing costs. The objective typically includes maximizing the vessel's capacity or minimizing its weight while adhering to design constraints such as maximum stress, deflection, and operating pressure. Design considerations may also encompass factors like thermal stresses, corrosion resistance, and manufacturability. Mathematical models, finite element analysis, and optimization algorithms are often utilized to iteratively refine designs and identify optimal solutions. The pressure vessel design problem plays a crucial role in ensuring the reliability and efficiency of industrial processes, with applications ranging from storage tanks and pipelines to aerospace propulsion systems. Effective solutions to this problem contribute to improved safety, reduced environmental impact, and enhanced operational performance in various engineering applications.

## 6. Result and Discussion

This section discusses the comparative results, runtime results, memory complexity, statistical analysis of M-IFDO against its competitive algorithms and balancing the exploration and exploitation of M-IFDO.

### 6.1. Comparative Results

At first glance, the standard test functions are used to evaluate the performance of M-IFDO. The results are examined against five well-known algorithms (IFDO, IMODE, HSES, LSHADE, and NBIPOP-aCMAES). Table (3) compares the results of the classical benchmark function between M-IFDO and its five competitive algorithms. The results show that the performance of M-IFDO in test functions TF (1, 3, 4, 6, 8, 9, 10, 11, 12, 15, 16, 17, 18 and 19) surpasses its counterpart algorithms. This superior performance is visually indicated by highlighting the corresponding values. Conversely, each of M-IFDO’s competitors has the highest performance in one benchmark problem as follows: IFDO, IMODE, HSES, LSHADE, and NBIPOP-aCMAES perform well in FT (14, 2, 5, 7, 13), respectively.

**Table (3): The comparative results of M-IFDO on classical benchmark functions with its counterpart algorithms**

| Test Functions | M-IFDO | | IFDO | | IMODE | | HSES | | LSHADE | | NBIPOP-aCMAES | |
|---|---|---|---|---|---|---|---|---|---|---|---|---|
| | **Mean** | **SD** | **Mean** | **SD** | **Mean** | **SD** | **Mean** | **SD** | **Mean** | **SD** | **Mean** | **SD** |
| **TF1** | **3.44E-24** | 1.12E-23 | 5.38E-24 | 2.74E-23 | 5.57E-24 | 2.45E-09 | 3.15E-17 | 6.56E-08 | 3.2E-51 | 2.11E-97 | 1.60E-02 | 2.41E-01 |
| **TF2** | 0.5172180 | 0.2712662 | 0.734345844 | 1.620259633 | **0.6893812** | 0.3029818 | 0.8159104 | 0.7197511 | 1.0391418 | 1.20039541 | 0.8140194 | 1.0242879 |
| **TF3** | **1.06E-13** | 3.93E-13 | 2.88E-07 | 6.90E-07 | 3.5522E-7 | 4.39552E-6 | 2.29E-07 | 2.1E-06 | 4.1298E-8 | 3.62324E-7 | 2.4971E-9 | 2.1564E-8 |
| **TF4** | **0.5E-04** | 0.0042682 | 3.60E-04 | 8.11E-04 | 7.388E-4 | 0.0034857 | 0.000998 | 0.001976 | 0.001719 | 0.003026 | 22.25302 | 2.301406 |
| **TF5** | 3.1E+01 | 40.019250 | 1.84E+02 | 3.45E+02 | 24.50103 | 60.3813791 | **6.900552** | 6.806474 | 64.21332 | 79.72729 | 163327.2 | 84,407.64 |
| **TF6** | **4.15E+06** | 1099.2393 | 4.22E+06 | 8.15E-09 | 1.522E-17 | 5.1461E-18 | 4.27E-16 | 1.42E-15 | 3.96E-17 | 1.48E-16 | 523.3889 | 238.5998 |
| **TF7** | 7.2E-01 | 0.3166318 | 5.68E-01 | 3.14E-01 | 0.494401 | 0.4251574 | 0.100291 | 0.005691 | **0.006073** | 0.004581 | 0.136871 | 0.062572 |
| **TF8** | **-3.00E+06** | 148152.30 | -2.92E+06 | 2.34E+05 | -2395203 | 210684.92 | -2757.53 | 393.6426 | -8.2E+11 | 1.4E+12 | -3307.28 | 154.5771 |
| **TF9** | **8.979103** | 9.84721 | 1.35E+01 | 6.66E+00 | 15.16542 | 4.702234 | 15.01823 | 3.15E+01 | 6.5788958 | 8.009807 | 24.91886 | 5.96036 |
| **TF10** | **3.891E-15** | 5.8771E-16 | 5.18E-15 | 1.67E-15 | 6.24E-15 | 2.3356822E-15 | 0.33101 | 0.567051 | 0.300138 | 0.551812 | 8.998715 | 2.001392 |
| **TF11** | **0.073453** | 0.039061 | 0.525690405 | 8.90E-02 | 0.498771 | 0.2042612 | 0.123304 | 0.063492 | 0.5465521 | 0.0772622 | 6.919952 | 3.92603 |
| **TF12** | **1.75E+01** | 18.610442 | 1.81E+01 | 2.57E+01 | 17.93835 | 25.174238 | 0.191105 | 0.088341 | 7.97E-14 | 2.41E-18 | 2058.501 | 4920.295 |
| **TF13** | 4.18E+09 | 3.1299918E7 | 4.10E+09 | 1.50E-05 | 11.0783 | 6.82023 | 58,037.21 | 88,136.71 | 0.002127 | 0.005643 | **0.102107** | 0.004643 |
| **TF14** | 8.5E-07 | 3.4511717E-5 | **2.68E-07** | 4.68E-07 | 3.8870E-7 | 6.1193E-7 | 104.712 | 90.64361 | 149.0016 | 134.4005 | 129.0993 | 20.92039 |
| **TF15** | **0.002E-16** | 6.7292255E-4 | 4.03E-16 | 9.25E-16 | 0.009504 | 0.0022439 | 192.0181 | 79.9039 | 178.8952 | 154.3831 | 115.0553 | 18.18352 |
| **TF16** | **1.94E-16** | 0.0334151 | 9.14E-16 | 3.61E-16 | 0.005372 | 0.0205601 | 448.9262 | 155.9724 | 213.2949 | 197.2351 | 363.9174 | 37.10534 |
| **TF17** | **2.20E+01** | 0.3212209 | 2.38E+01 | 1.24E-01 | 22.72014 | 0.3149435 | 576.6621 | 161.0611 | 456.5449 | 170.9483 | 504.8481 | 34.19476 |
| **TF18** | **2.23E+02** | 0.0133942 | 2.24E+02 | 2.68E-05 | 223.9612 | 9.1675E-6 | 231.1516 | 177.9091 | 138.2759 | 159.0189 | 119.638 | 53.10185 |
| **TF19** | **3.15E+01** | 0.0789975 | 3.15E+01 | 1.32E-03 | 23.7201 | 0.0303582 | 689.584 | 189.9014 | 791.6331 | 207.1292 | 554.1038 | 14.20162 |

Secondly, the results of the CEC-C06 2019 benchmarking functions for the M-IFDO and its competitive algorithms are shown in Table (4). For each test function in the earlier mentioned table, superior results are shown in bold. The outcomes of the M-IFDO algorithm in the CEC (02 and 09) tests are almost the same as those obtained by the IFDO method. On the other hand, M-IFDO outperforms alternative rival algorithms in tests conducted by CEC (01, 03, 04, 05, 07, 08, and 10). Also, IMODE is the best performance algorithm for CEC06.

Table (4): The comparative results of M-IFDO on CEC-C06 2019 benchmark functions with its counterpart algorithms

| Test Functions | M-IFDO | | IFDO | | IMODE | | HSES | | LSHADE | | NBIPOP-aCMAES | |
|---|---|---|---|---|---|---|---|---|---|---|---|---|
| | Mean | SD | Mean | SD | Mean | SD | Mean | SD | Mean | SD | Mean | SD |
| **CEC01** | **2.65E+03** | 1.39E+04 | 5.08E+08 | 3.49E+08 | 4.49E+03 | 3.00E+04 | 5.60E+04 | 6.99E+04 | 4.62E+04 | 5.69E+04 | 6.46E+04 | 4.93E+04 |
| **CEC02** | **3.9011** | 0.02113 | **4.000002** | 1.00E-05 | 4.2324 | 3.52414E-9 | 77.0368 | 81.7888 | 14.3495 | 0.00456 | 18.3434 | 0.0006 |
| **CEC03** | **13.7024** | 3.50993E-5 | 13.7024 | 4.82E-09 | 12.7023 | 1.8490E-11 | 12.7026 | 0.00071 | 12.9024 | 0.0 | 12.9025 | 0.00032 |
| **CEC04** | **3.12E+01** | 1.29E+01 | 3.98E+01 | 1.06E+01 | 3.61E+01 | 1.45E+01 | 3.64E+02 | 4.24E+02 | 4.05E+02 | 2.59E+02 | 4.27E+01 | 2.32E+01 |
| **CEC05** | **1.13E+00** | 7.06E-02 | 1.15E+00 | 7.66E-02 | 2.24E+00 | 7.67E-02 | 2.55E+00 | 3.26E-01 | 2.63E+00 | 2.82E-01 | 2.31E+00 | 1.16E-01 |
| **CEC06** | 1.21E+01 | 5.21E-01 | 1.31E+01 | 6.00E-01 | **1.11E+01** | 8.05E-01 | 9.80E+00 | 1.54E+00 | 1.08E+01 | 1.04E+00 | 6.07E+00 | 1.48E+00 |
| **CEC07** | **1.36E+01** | 5.79E+02 | 1.21E+02 | 1.35E+01 | 1.20E+02 | 1.16E+02 | 1.03E+01 | 3.29E+02 | 4.91E+02 | 1.95E+02 | 4.10E+02 | 2.91E+02 |
| **CEC08** | **4.24E+00** | 8.29E-01 | 4.94E+00 | 8.81E-01 | 5.10E+00 | 6.57E-01 | 6.67E+00 | 5.22E-01 | 6.81E+00 | 4.17E-01 | 5.37E+00 | 5.46E-01 |
| **CEC09** | **2.0** | 5.54501E-4 | **2.0** | 3.10E-15 | 5.8371 | 1.5566 | 6.1467 | 2.871 | **2.0** | 1.4916E-10 | 3.5704 | 0.3362 |
| **CEC10** | **1.91828** | 4.44089E-16 | 2.7182818 | 4.54E-15 | 2.3181 | 8.7817E-16 | 19.2604 | 0.1315 | 21.4761 | 0.2111 | 20.04 | 0.068 |

### 6.2. Runtime Results

The execution time of M-IFDO on the classical testing functions is compared with its counterpart algorithms. Table (5) presents the outcomes of the total time in brief. The results of traditional benchmark test functions depict that M-IFDO takes less time to test functions in TF1, TF2, TF4, TF7, TF8, TF9, TF11, TF12, TF13, TF14, TF15, TF16, TF17, TF18 and TF19. Besides, IFDO, IMODE, and LSHADE require less execution time for TF3, TF6 and TF10, respectively.

**Table (5): Outcomes of M-IFDO and IFDO Execution Time for Classical Benchmark Test Functions**

| Test functions | M-IFDO | IFDO | IMODE | HSES | LSHADE | NBIPOP-aCMAES |
|---|---|---|---|---|---|---|
| **TF1** | **12.185795** | 14.247080 | 14.137993 | 16.546220 | 17.794322 | 14.16038 |
| **TF2** | **11.015131** | 13.269279 | 18.225728 | 16.047039 | 12.477624 | 11.76668 |
| **TF3** | 12.011056 | **11.709830** | 14.596870 | 12.445491 | 13.989193 | 18.56795 |
| **TF4** | **10.884500** | 12.278314 | 13.517973 | 16.450461 | 15.405507 | 19.88174 |
| **TF5** | 8.882539 | 15.540126 | 19.852576 | 10.083041 | 6.864801 | 21.73414 |
| **TF6** | 14.530318 | 16.465144 | **13.957799** | 17.807292 | 15.588198 | 17.43181 |
| **TF7** | **5.028262** | 9.257813 | 13.745315 | 11.588522 | 12.200169 | 20.52254 |
| **TF8** | **17.993566** | 18.664209 | 25.395976 | 19.518810 | 20.021920 | 20.42676 |
| **TF9** | **17.938539** | 18.688155 | 19.425239 | 20.883634 | 18.329175 | 22.62755 |
| **TF10** | 11.018423 | 13.612513 | 18.046302 | 16.892365 | **11.609797** | 19.98769 |
| **TF11** | **14.377516** | 16.538066 | 17.926955 | 18.689947 | 16.247844 | 19.27857 |
| **TF12** | **15.453290** | 15.921168 | 17.032557 | 16.558051 | 17.029514 | 18.582 |
| **TF13** | **9.765265** | 13.370683 | 16.772819 | 12.649700 | 12.363544 | 12.53924 |
| **TF14** | **12.301826** | 14.109915 | 18.227620 | 15.679479 | 14.254859 | 19.54956 |
| **TF15** | **10.373158** | 12.416894 | 12.336857 | 19.057239 | 15.440824 | 12.02551 |
| **TF16** | **11.191316** | 12.823273 | 16.015560 | 14.260564 | 13.498449 | 15.94858 |
| **TF17** | **14.410829** | 15.228403 | 18.761425 | 16.726982 | 18.302501 | 16.34374 |
| **TF18** | **16.326270** | 19.717039 | 17.151084 | 21.882230 | 23.481757 | 24.02835 |
| **TF19** | **19.568461** | 19.656789 | 23.136551 | 21.903008 | 22.476079 | 23.03601 |

Meanwhile, the execution time of M-IFDO on the IEEE CEC 2019 benchmark functions is compared with its counterpart algorithms. Table (6) shows the outcomes of the total time in brief. From the outcomes, the execution time of modern CEC benchmark 2019 test functions results for the M-IFDO algorithm is better than the other algorithms in CEC01, CEC03, CEC04, CEC05, CEC07, CEC08 and CEC09. In the CEC02 test function, IFDO needs less execution time than the other methods. Likewise, IMODE requires less time to solve the CEC06 test function and NBIPOP-aCMAES takes less time than the others in CEC10. Overall, M-IFDO needs less time than the other algorithms to solve the CEC benchmark 2019.

**Table (6): Outcomes of M-IFDO and IFDO Execution Time for IEEE CEC Benchmark 2019**

| Test functions | M-IFDO | IFDO | IMODE | HSES | LSHADE | NBIPOP-aCMAES |
|---|---|---|---|---|---|---|
| **CEC01** | **14.617488** | 17.433333 | 39.165498 | 24.223746 | 34.982767 | 32.591619 |
| **CEC02** | 22.869618 | **21.164457** | 32.955562 | 36.310162 | 35.790281 | 34.687751 |
| **CEC03** | **19.977707** | 22.080011 | 31.363230 | 31.917644 | 28.758459 | 25.024420 |
| **CEC04** | **17.513396** | 24.396022 | 20.544143 | 24.910597 | 31.369401 | 24.146576 |
| **CEC05** | **15.228305** | 20.607275 | 31.906933 | 38.955995 | 23.949873 | 23.005748 |
| **CEC06** | 24.700089 | 26.064056 | **22.594776** | 26.059386 | 29.862689 | 32.481352 |
| **CEC07** | **26.721411** | 27.749142 | 28.426762 | 31.894016 | 28.471136 | 34.841033 |
| **CEC08** | **23.556675** | 32.894285 | 33.563744 | 34.125571 | 28.764646 | 34.665611 |
| **CEC09** | **22.393615** | 31.394046 | 30.579181 | 35.349765 | 24.053061 | 31.572071 |
| **CEC10** | 26.266701 | 28.967531 | 28.914464 | 31.804724 | 33.528819 | **24.288242** |

Finally, the outcomes of Table (7), present the execution time of five real-world applications. The M-IFDO has more ability compared to the IFDO from the execution time on the AAAD, TSP, JSP, and PVDP applications. Conversely, IFDO needs less time to solve COP compared with its counterpart algorithms.

**Table (7): Execution time of M-IFDO on five real-world applications compared with its counterpart algorithms.**

| Application name | M-IFDO | IFDO | IMODE | HSES | LSHADE | NBIPOP-aCMAES |
|---|---|---|---|---|---|---|
| **AAAD** | **13.484731** | 17.948202 | 21.009383 | 26.001874 | 15.028173 | 19.0018471 |
| **TSP** | **16.018349** | 18.001937 | 19.381363 | 20.118371 | 21.814401 | 17.1938445 |
| **COP** | 22.119048 | **21.038474** | 23.001832 | 22.981743 | 27.209734 | 29.1840562 |
| **JSP** | **24.771946** | 27.104857 | 28.039471 | 25.184766 | 31.004975 | 30.3746137 |
| **PVDP** | **30.229571** | 31.494851 | 32.984746 | 33.948471 | 32.880047 | 30.9284741 |

### 6.3. Memory Complexity

This section analyses the overall performance benchmarking of M-IFDO (memory consumption) based on classical functions and CEC-C06 2019. Table (8) presents memory consumption for the 30 solutions obtained by M-IFDO and its competitive algorithms for classical benchmark functions. We observe that M-IFDO consumes less memory than the other techniques in 14 test functions out of 19. Surprisingly, IFDO requires minimum memory usage for FT5 and FT18. Simultaneously, IMODE necessitates minimal memory utilization for FT7 and FT13. In general, the proposed M-IFDO consumes less memory than the other methods.

**Table (8): Average running time with memory consumption for classical benchmark functions for the 30 solutions obtained by M-IFDO with its counterpart algorithms.**

| Test Functions | M-IFDO | IFDO | IMODE | HSES | LSHADE | NBIPOP-aCMAES | Winner |
|---|---|---|---|---|---|---|---|
| **TF1** | **44.231** | 65.016 | 52.480 | 109.780 | 68.144 | 70.145 | M-IFDO |
| **TF2** | 95.112 | 108.213 | 110.436 | 125.095 | **93.411** | 110.691 | LSHADE |
| **TF3** | **82.112** | 135.985 | 96.783 | 111.420 | 90.972 | 89.123 | M-IFDO |
| **TF4** | **70.654** | 116.659 | 119.129 | 81.478 | 71.498 | 78.807 | M-IFDO |
| **TF5** | 65.167 | **56.570** | 143.676 | 132.381 | 97.307 | 143.076 | IFDO |
| **TF6** | **49.624** | 103.296 | 81.952 | 71.230 | 57.382 | 102.840 | M-IFDO |
| **TF7** | 52.389 | 132.137 | **43.400** | 68.942 | 148.580 | 128.028 | IMODE |
| **TF8** | **51.251** | 85.843 | 92.792 | 52.841 | 76.450 | 102.189 | M-IFDO |
| **TF9** | **63.905** | 92.183 | 68.434 | 97.501 | 72.976 | 96.628 | M-IFDO |
| **TF10** | **44.516** | 47.940 | 84.732 | 76.536 | 81.730 | 79.263 | M-IFDO |
| **TF11** | **70.301** | 115.990 | 97.076 | 78.906 | 116.173 | 109.086 | M-IFDO |
| **TF12** | **47.117** | 56.308 | 79.884 | 92.018 | 95.314 | 74.092 | M-IFDO |
| **TF13** | 52.482 | 61.119 | **50.108** | 77.048 | 85.677 | 79.123 | IMODE |
| **TF14** | **103.062** | 108.455 | 121.230 | 142.408 | 101.721 | 158.105 | M-IFDO |
| **TF15** | **72.897** | 94.026 | 82.186 | 105.216 | 102.647 | 80.978 | M-IFDO |
| **TF16** | **91.462** | 92.869 | 98.885 | 141.031 | 105.784 | 125.276 | M-IFDO |
| **TF17** | **81.715** | 108.799 | 133.457 | 101.145 | 79.768 | 117.929 | M-IFDO |
| **TF18** | 95.856 | **81.474** | 107.093 | 84.354 | 133.713 | 86.839 | IFDO |
| **TF19** | **78.417** | 145.459 | 93.244 | 145.329 | 100.072 | 89.261 | M-IFDO |
| **Total winners** | 14 | 2 | 2 | 0 | 1 | 0 | |

Moreover, Table (9) presents the average running time with memory consumption for IEEE CEC Benchmark 2019 for the 30 solutions obtained by M-IFDO with its counterpart algorithms. The results show that M-IFDO consumes less memory than the other techniques in 6 CEC out of 9. Surprisingly, IFDO, IMODE, and NBIPOP-aCMAES require minimum memory usage for CEC05, CEC03, and CEC04, respectively. In general, the proposed M-IFDO consumes less memory than the other methods for IEEE CEC Benchmark 2019.

**Table (9): Average running time with memory consumption for IEEE CEC Benchmark 2019 for the 30 solutions obtained by M-IFDO with its counterpart algorithms.**

| **Test Functions** | **M-IFDO** | **IFDO** | **IMODE** | **HSES** | **LSHADE** | **NBIPOP-aCMAES** | **Winner** |
|---|---|---|---|---|---|---|---|
| **CEC01** | **119.408** | 120.208 | 132.578 | 137.628 | 95.263 | 191.458 | M-IFDO |
| **CEC02** | **139.034** | 165.887 | 184.534 | 164.281 | 110.349 | 146.484 | M-IFDO |
| **CEC03** | 137.638 | 142.529 | **112.898** | 117.714 | 128.302 | 196.485 | IMODE |
| **CEC04** | 107.756 | 168.340 | 157.683 | 104.773 | 172.005 | **101.736** | NBIPOP-aCMAES |
| **CEC05** | 124.116 | **121.806** | 163.761 | 181.922 | 107.831 | 144.460 | IFDO |
| **CEC06** | **111.147** | 111.363 | 192.043 | 115.822 | 147.743 | 171.598 | M-IFDO |
| **CEC07** | **105.995** | 189.281 | 121.882 | 146.234 | 106.121 | 169.953 | M-IFDO |
| **CEC08** | **129.952** | 133.814 | 167.854 | 157.958 | 152.684 | 129.868 | M-IFDO |
| **CEC09** | **148.365** | 159.137 | 191.440 | 165.166 | 157.645 | 159.211 | M-IFDO |
| **Total winners** | 6 | 1 | 1 | 0 | 0 | 1 | |

Lastly, based on the results shown in Table (10) about the memory utilization of M-IFDO on five real-world applications. It can be observed that the M-IFDO outperforms the IFDO in terms of memory usage on the AAAD, TSP, JSP, and PVDP applications. On the other hand, COP needs less memory using IFDO compared with its competitive algorithms.

**Table (10): Memory consumption of M-IFDO on five real-world applications compared with its counterpart algorithms.**

| Application name | M-IFDO | IFDO | IMODE | HSES | LSHADE | NBIPOP-aCMAES |
|---|---|---|---|---|---|---|
| **AAAD** | **119.963** | 128.005 | 148.919 | 132.938 | 134.673 | 127.836 |
| **TSP** | **98.756** | 101.280 | 104.839 | 109.018 | 110.009 | 107.128 |
| **COP** | 113.018 | 117.282 | **112.183** | 117.189 | 118.129 | 120.091 |
| **JSP** | **121.018** | 124.192 | 125.122 | 122.129 | 129.005 | 131.237 |
| **PVDP** | **109.126** | 111.038 | 117.394 | 116.917 | 115.028 | 110.381 |

### 6.4. Statistical Analysis

At times, optimization algorithms may yield both the optimal and suboptimal solutions for a given problem. For instance, if an algorithm is run twice on a certain problem, it may produce the best solution initially and the worst one thereafter, or vice versa. The literature utilized statistical tools [60] to compare the efficacy of M-IFDO in problem-solving with other evolutionary algorithms, determining their success or failure. Our experiment utilized seven statistical measures to address numerical optimization problems: mean, standard deviation, best, worst, average computing time, number of successful minimizations, and number of failure minimizations. Pairwise statistical testing tools such as the Wilcoxon signed-rank test can be utilised to compare two algorithms and determine which algorithm can solve a specific optimisation problem with higher statistical success [60][61]. In the experiment, M-IFDO was compared with other algorithms by using the Wilcoxon signed-rank test, where the significant statistical value (α) was considered as 0.05, and the null hypothesis (H0) for a specific benchmark problem was defined in Equation (15) as follows:

$$Median\ (Algorithm\ A) = Median\ (Algorithm\ B) \qquad (15)$$

To determine the algorithm that achieved a better solution in terms of statistics, T+, T-, and p-value were provided by the Wilcoxon signed-rank test to determine the ranking size. In the same experiment, GraphPad Prism was used to determine the T+ and T- values that ranged from 0 to 465. Table (10) presents a statistical comparison to find the algorithm that provides the best solution for the solved classical benchmark functions using two-sided Wilcoxon Signed-Rank Test (α=0.05)

**Table (10): Statistical comparison for the classical benchmark functions using two-sided Wilcoxon Signed-Rank Test (α=0.05)**

| Test Functions | M-IFDO vs IFDO | | | | M-IFDO vs IMODE | | | | M-IFDO vs HSES | | | | M-IFDO vs LSHADE | | | | M-IFDO vs NBIPOP-aCMAES | | | |
|---|---|---|---|---|---|---|---|---|---|---|---|---|---|---|---|---|---|---|---|---|
| | p-value | T+ | T- | Win | p-value | T+ | T- | Win | p-value | T+ | T- | Win | p-value | T+ | T- | Win | p-value | T+ | T- | Win |
| **TF1** | 0.0001 | 40 | 425 | + | 0.5322 | 154 | 311 | - | 0.0002 | 196 | 269 | + | 0.719 | 182 | 283 | - | 0.0005 | 323 | 142 | + |
| **TF2** | 0.0002 | 445 | 20 | + | 0.0009 | 77 | 388 | + | 0.0008 | 422 | 43 | + | 0.0505 | 435 | 30 | + | 0.5233 | 136 | 329 | - |
| **TF3** | <0.0001 | 50 | 415 | + | 0.0005 | 260 | 205 | + | 1.0000 | 0 | 0 | = | 0.0003 | 410 | 55 | + | 0.0002 | 303 | 162 | + |
| **TF4** | 0.0316 | 296 | 169 | + | 1.0000 | 0 | 0 | = | 0.0027 | 456 | 9 | + | 0.0009 | 202 | 263 | + | 0.1538 | 386 | 79 | + |
| **TF5** | 1.0000 | 0 | 0 | = | 0.479 | 36 | 429 | + | 0.0009 | 86 | 379 | + | 0.0001 | 411 | 54 | + | 0.0007 | 362 | 103 | + |
| **TF6** | 0.0008 | 311 | 154 | + | <0.0001 | 369 | 96 | + | 0.005 | 122 | 343 | + | 0.0415 | 269 | 196 | + | 0.0005 | 247 | 218 | + |
| **TF7** | 0.0005 | 172 | 293 | + | 0.0007 | 227 | 238 | + | <0.0001 | 216 | 249 | + | 0.0713 | 10 | 455 | + | 0.0006 | 328 | 137 | + |
| **TF8** | 0.0008 | 220 | 245 | + | 0.35 | 456 | 9 | + | 0.3093 | 260 | 205 | + | 0.0348 | 361 | 104 | + | 0.0008 | 23 | 442 | + |
| **TF9** | 0.0002 | 120 | 345 | + | 0.0003 | 459 | 6 | + | 0.0432 | 67 | 398 | + | 1.000 | 0 | 0 | = | 0.1082 | 30 | 435 | + |
| **TF10** | 0.0311 | 301 | 164 | + | 0.0009 | 212 | 253 | + | 0.3018 | 299 | 166 | + | 0.0409 | 21 | 444 | + | 0.0323 | 385 | 80 | + |
| **TF11** | 1.0000 | 0 | 0 | = | 0.507 | 197 | 268 | - | 0.5282 | 61 | 404 | - | 0.5157 | 29 | 436 | - | <0.0001 | 199 | 266 | + |
| **TF12** | 0.0002 | 24 | 441 | + | 0.0009 | 201 | 264 | + | 0.0087 | 349 | 116 | + | 0.0091 | 133 | 332 | + | 0.3025 | 341 | 124 | + |
| **TF13** | <0.0001 | 152 | 313 | + | 0.0001 | 373 | 92 | + | 0.0007 | 259 | 206 | + | 0.1157 | 287 | 178 | + | 0.0002 | 106 | 359 | + |
| **TF14** | 0.4771 | 29 | 436 | + | 0.0002 | 169 | 296 | + | 0.3739 | 378 | 87 | + | 0.0085 | 293 | 172 | + | 0.0333 | 240 | 225 | + |
| **FT15** | 0.0258 | 272 | 193 | - | 0.0003 | 195 | 270 | + | 0.1227 | 101 | 364 | + | 0.0002 | 444 | 21 | + | 0.3496 | 319 | 146 | + |
| **TF16** | 1.0000 | 0 | 0 | = | 0.0009 | 456 | 9 | + | 0.4352 | 65 | 400 | + | 0.2897 | 274 | 191 | + | 0.0004 | 394 | 71 | + |
| **TF17** | 0.0008 | 198 | 267 | + | <0.0001 | 183 | 282 | + | 0.4312 | 192 | 273 | + | 0.4546 | 133 | 332 | + | 0.0203 | 194 | 271 | + |
| **TF18** | 0.5133 | 81 | 384 | - | 0.0003 | 125 | 340 | + | 0.0003 | 8 | 457 | + | 0.3393 | 168 | 297 | + | 0.5324 | 280 | 185 | - |
| **TF19** | 0.0001 | 75 | 390 | + | 0.0008 | 137 | 328 | + | 0.0004 | 427 | 38 | + | 0.0713 | 167 | 298 | + | 0.0008 | 40 | 425 | + |
| **+/=/-** | | | | **14/3/2** | | | | **16/1/2** | | | | **17/1/1** | | | | **16/1/2** | | | | **17/0/2** |

Table (11) presents a statistical comparison to find the algorithm that provides the best solution for the solved CEC-C06 2019 benchmark functions using the two-sided Wilcoxon Signed-Rank Test (α=0.05).

**Table (11): Statistical comparison for the CEC-C06 2019 benchmark functions using two-sided Wilcoxon Signed-Rank Test (α=0.05)**

| Test Functions | M-IFDO vs IFDO | | | | M-IFDO vs IMODE | | | | M-IFDO vs HSES | | | | M-IFDO vs LSHADE | | | | M-IFDO vs NBIPOP-aCMAES | | | |
|---|---|---|---|---|---|---|---|---|---|---|---|---|---|---|---|---|---|---|---|---|
| | p-value | T+ | T- | Win | p-value | T+ | T- | Win | p-value | T+ | T- | Win | p-value | T+ | T- | Win | p-value | T+ | T- | Win |
| **CEC01** | 0.0003 | 395 | 70 | + | 0.0002 | 416 | 49 | + | 0.5105 | 430 | 35 | - | 0.0003 | 103 | 362 | + | 0.0003 | 133 | 332 | + |
| **CEC02** | 1.0000 | 0 | 0 | = | 0.5587 | 322 | 143 | - | 0.0007 | 350 | 115 | + | 1.0000 | 0 | 0 | = | 0.0007 | 195 | 270 | + |
| **CEC03** | 0.0008 | 256 | 209 | + | 0.0009 | 110 | 355 | + | 0.0005 | 432 | 33 | + | 1.0000 | 0 | 0 | = | 0.0001 | 50 | 415 | + |
| **CEC04** | 0.002 | 369 | 96 | + | 1.0000 | 0 | 0 | = | 0.0003 | 224 | 241 | + | 0.0014 | 219 | 246 | + | 0.0005 | 383 | 82 | + |
| **CEC05** | 0.5785 | 215 | 250 | - | 0.0005 | 269 | 196 | + | 0.0001 | 246 | 219 | + | 0.0002 | 94 | 371 | + | 0.0004 | 189 | 276 | + |
| **CEC06** | 0.0006 | 171 | 294 | + | 0.0003 | 81 | 384 | + | 0.0005 | 420 | 45 | + | 0.0003 | 217 | 248 | + | 0.0002 | 151 | 314 | + |
| **CEC07** | <0.0001 | 119 | 346 | + | 0.0007 | 119 | 346 | + | 0.0001 | 171 | 294 | + | 0.0006 | 384 | 81 | + | <0.0001 | 314 | 151 | + |
| **CEC08** | 0.0002 | 149 | 316 | + | 0.0003 | 62 | 403 | + | 0.0004 | 227 | 238 | + | <0.0001 | 66 | 399 | + | 0.5072 | 354 | 111 | - |
| **CEC09** | 0.0008 | 371 | 94 | + | 0.0006 | 66 | 399 | + | 0.0002 | 87 | 378 | + | 0.0005 | 219 | 246 | + | 0.0001 | 130 | 335 | + |
| **CEC10** | 0.0001 | 107 | 358 | + | 0.0002 | 172 | 293 | + | <0.0001 | 144 | 321 | + | 0.0002 | 214 | 251 | + | 0.5009 | 250 | 215 | - |
| **+/=/-** | | | | **7/1/1** | | | | **7/1/1** | | | | **8/0/1** | | | | **7/2/0** | | | | **7/0/2** |

Meanwhile, M-IFDO exhibits several limitations within its operational scope. Firstly, its sensitivity to initial parameters can significantly impact its performance, potentially leading to slower convergence or premature convergence to suboptimal solutions. Moreover, the effectiveness of M-IFDO is intricately tied to the population size or number of agents utilized; inadequate agent numbers might hinder proper exploration of the search space, compromising solution quality. Additionally, M-IFDO’s computational complexity escalates notably when tackling intricate, high-dimensional problems, consequently increasing the computational cost and time required for convergence. Its adaptability in dynamically changing environments or when encountering abrupt shifts in the search landscape might also be limited. Furthermore, constraints pose a challenge for M-IFDO in ensuring feasibility while optimizing within constraint-bound scenarios. Lastly, M-IFDO might struggle with non-smooth, discontinuous, or irregular objective functions, impeding its ability to effectively navigate and converge to optimal solutions within such function landscapes. Overcoming these limitations demands ongoing research focusing on algorithmic enhancements, refined parameter tuning strategies, adaptive mechanisms, and specialized techniques tailored to address specific problem types, thus aiming to augment M-IFDO’s performance across diverse application domains.

### 6.5. Balancing Exploration and Exploitation of M-IFDO

M-IFDO adeptly balances the dual objectives of exploration and exploitation to navigate complex solution spaces in search of optimal solutions. Exploration involves broadly searching through the solution space to discover diverse potential solutions and avoid premature convergence to suboptimal local optima, ensuring the algorithm can find the global optimum. Exploitation, on the other hand, focuses on intensively searching within the most promising areas identified during exploration, refining these solutions to improve their quality. In M-IFDO, fitness-based position updates play a crucial role in managing this balance, with agents' positions updated based on their fitness levels—initially promoting exploration as agents are scattered, then gradually favoring exploitation as they converge towards higher fitness regions. To maintain robust exploration, M-IFDO incorporates random walks into its position updates, introducing stochasticity to explore new regions continually, which is particularly vital in early optimization stages. Additionally, adaptive mechanisms adjust the balance of exploration and exploitation over time, initially favoring exploration and shifting towards exploitation as more promising regions are identified. Hybrid strategies, combining exploration and exploitation in different phases or iterations, further ensure a dynamic and flexible search process. The success of M-IFDO depends on well-designed fitness functions, careful parameter tuning (such as the size of random walks, fitness influence, and convergence rate), and maintaining diversity among agents to prevent premature convergence. Visualizing these dynamics, charts illustrate the balance of exploration and exploitation over iterations, the progression of fitness values indicating convergence to better solutions, and the diversity of solutions maintained to avoid premature convergence. In a nutshell, the M-IFDO's effectiveness lies in its dynamic adjustment of exploration and exploitation, ensuring comprehensive search and fine-tuning of solutions to achieve optimal results. . Figure 3 depicts the balancing of exploration and exploitation of M-IFDO over time.

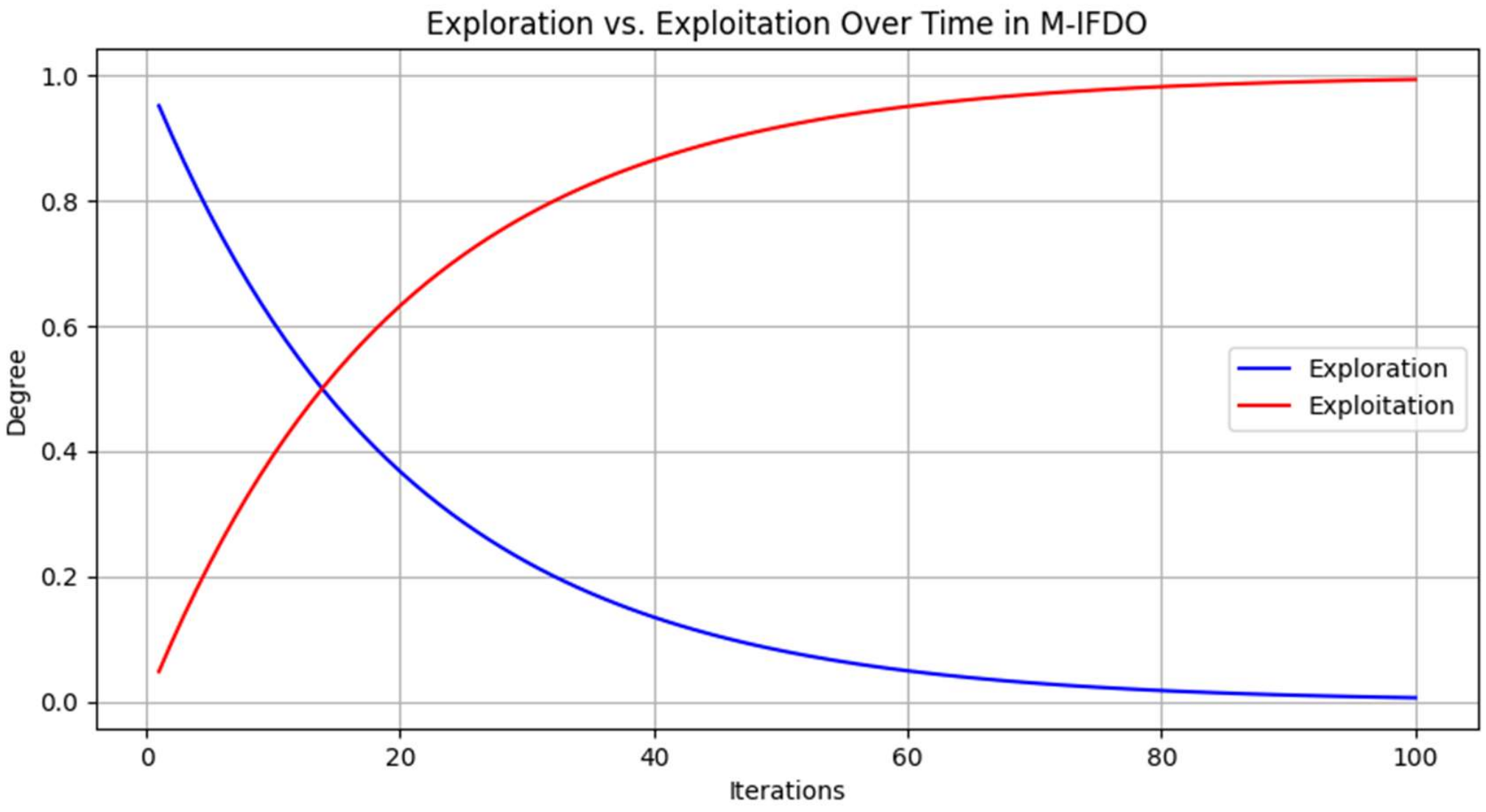

**Figure 3: Exploration and exploitation of M-IFDO over time**

### 6.6. Diversity Analysis of M-IFDO

Diversity analysis in the M-IFDO is crucial for ensuring a comprehensive search of the solution space and preventing premature convergence to local optima. This analysis involves assessing how well the algorithm maintains a variety of candidate solutions throughout the optimization process. The M-FDO algorithm employs several mechanisms to preserve this diversity. Fitness-based position updates ensure that agents with higher fitness influence the movement of other agents, promoting exploration when agents are scattered and shifting towards exploitation as they converge. Random walks introduce stochasticity into position updates, allowing the algorithm to explore new regions of the solution space continually. Adaptive mechanisms adjust the balance between exploration and exploitation over time, favoring exploration in the early stages and exploitation as more promising regions are identified. Techniques such as crowding and niching are also used to maintain diversity by preventing too many solutions from converging to the same region, ensuring multiple peaks in the fitness landscape are explored. High diversity is essential in the early stages to avoid local optima, while a controlled reduction in diversity later allows for the fine-tuning of the best solutions. Overall, effective diversity management in M-FDO ensures a robust and flexible search process, capable of solving complex optimization problems efficiently.

## 7. Conclusion

This paper introduced a modified version of IFDO, referred to as M-IFDO. The improvement was achieved by relocating the scout bees to the IFDO to facilitate the movement of the scout bees and attain superior performance and an optimal solution. To be more precise, two characteristics, namely alignment and cohesiveness, were eliminated from IFDO. Instead, the Lambda parameter was substituted for alignment and cohesion. To assess the effectiveness of the recently implemented M-IFDO algorithm, it was subjected to testing on a total of 19 fundamental benchmark functions, as well as 10 problems from the IEEE Congress of Evolutionary Computation (CEC-C06 2019) and a real-world scenario. M-IFDO was evaluated against five cutting-edge algorithms. The verification criteria are determined by the algorithm's convergence, memory utilization, and statistical outcomes. The experimental results indicate that M-IFDO outperforms its competitors in multiple instances on both the benchmark functions and the engineering problem. In general, M-IFDO exhibited superior performance across the majority of classical and CEC-C06 2019 benchmark functions. Meanwhile, the results showed that M-IFDO had a shorter solving time for the majority of the testing functions and four real-world applications. Furthermore, the M-IFDO approach has a lower memory consumption compared to other methods for both classical and CEC-C06 2019 benchmark functions, and four real-world problems. Regarding statistical evaluation, the suggested method demonstrated superior performance in solving classical and CEC-C06 2019 benchmark functions. This was determined using a two-sided Wilcoxon Signed-Rank Test. However, M-IFDO may struggle to traverse and converge to optimal solutions in non-smooth, discontinuous, or irregular objective functions. To further improve M-IFDO's performance across varied application domains, ongoing research must focus on algorithmic upgrades, enhanced parameter tuning strategies, adaptive mechanisms, and specialized techniques for certain problem types. We have also noticed that in recent years, FDO is still considered one of the most promising optimization algorithms in terms of being integrated with other technologies. This means that the changes and developments in it are continuing to come up with new methods that can be more effective and capable of yielding more satisfying results. For future reading, the authors advise the reader could optionally read the following research works [62–79].

**Acknowledgments**

The authors wish to express sincere thanks to their universities for providing facilities and continuous support in conducting this study.

**Compliance with Ethical Standards**

**Conflict of interest:** Not declared.

**Funding:** Still waiting to be received.

**Competing Interests:** Authors declare that there are not any competing interests.

**Ethical and informed consent for data used:** Not applicable.

**Data availability and access:** Data supporting this study are included within the article.